\documentclass[letterpaper]{article} 
\usepackage {aaai23}  
\usepackage{times}  
\usepackage{helvet}  
\usepackage{courier}  
\usepackage[hyphens]{url}  
\usepackage{graphicx} 
\def\UrlFont{\rm}  
\usepackage{natbib}  
\usepackage{caption} 
\usepackage{comment} 
\usepackage[T1]{fontenc}
\usepackage{algorithm}
\usepackage{algorithmic}

\usepackage{newfloat}
\usepackage{listings}
\DeclareCaptionStyle{ruled}{labelfont=normalfont,labelsep=colon,strut=off} 
\floatstyle{ruled}
\newfloat{listing}{tb}{lst}{}
\floatname{listing}{Listing}
\title{MATT: Multimodal Attention Level Estimation for e-learning Platforms}
\author{
    Roberto Daza,
    Luis F. Gomez,
    Aythami Morales,
    Julian Fierrez,
    Ruben Tolosana,
    Ruth Cobos,
    Javier Ortega-Garcia\\
}
\affiliations{
    School of Engineering, Autonomous University of Madrid\\
    \{roberto.daza, luisf.gomez, aythami.morales, julian.fierrez, ruben.tolosana, ruth.cobos, javier.ortega\}@uam.es
   
}

\usepackage{bibentry}

\begin{document}

\maketitle

\begin{abstract}
This work presents a new multimodal system for remote attention level estimation based on multimodal face analysis. Our multimodal approach uses different parameters and signals obtained from the behavior and physiological processes that have been related to modeling cognitive load such as faces gestures (e.g., blink rate, facial actions units) and user actions (e.g., head pose, distance to the camera).  The multimodal system uses the following modules based on Convolutional Neural Networks (CNNs): Eye blink detection, head pose estimation, facial landmark detection, and facial expression features. First, we individually evaluate the proposed modules in the task of estimating the student’s attention level captured during online e-learning sessions. For that we trained binary classifiers (high or low attention) based on Support Vector Machines (SVM) for each module. Secondly, we find out to what extent multimodal score level fusion improves the attention level estimation. The mEBAL database is used in the experimental framework, a public multi-modal database for attention level estimation obtained in an e-learning environment that contains data from 38 users while conducting several e-learning tasks of variable difficulty (creating changes in student cognitive loads).
\end{abstract}

\section{INTRODUCTION}

Over the last years, new technologies have brought a substantial change enabling many processes and applications to move from the physical to the digital world. Education has been one of the most affected areas, as remote e-learning offers many advantages over traditional learning (e.g., during the COVID-19 outbreak). Currently, e-learning and virtual education platforms are a fundamental pillar in the improvement strategy of the most important academic institutions such as Stanford, Oxford, Harvard, etc. Future looks bright for the e-learning industry, with remarkable growth numbers in the recent years and estimating even better results for the next 10 years \cite{chen2018research}.

E-learning presents many advantages \cite{bowers2015students}: Flexible schedules, allows a higher number of students, etc; however, it also presents challenges in comparison with the traditional face-to-face education system. 
Without a doubt, the recent e-learning platforms \cite{hernandez2019edbb} are a key tool to overcome these difficulties. These platforms allow the monitoring of e-learning sessions, capturing student’s information for a better understanding of the student’s behaviors and conditions. These platforms may incorporate new technologies to analyze students' information such as the attention level \cite{daza2021alebk}, the heart rate \cite{hernandez2020heart}, the emotional state \cite{shen2009affective}, and the gaze and head pose \cite{asteriadis2009estimation}.

E-learning platforms represent indeed an opportunity to improve education. In the traditional face-to-face education system, how can a teacher know when students present higher or lower attention levels? In remote education this is even more difficult to infer without direct contact between the teacher and the student. On the other hand, the automatic estimation of the student’s attention level on e-learning platforms is feasible \cite {pengpredicting}, and represents a high value tool to improve face-to-face and online education. 

This information obtained from e-learning platforms can be used to create personalized environments and a more secure evaluation, for example, to: i) adapt dynamically the environment and content \cite{nappi2018context, 2018_INFFUS_MCSreview1_Fierrez} based on the attention level of the students, and ii) improve the educational materials and resources with a further analysis of the e-learning sessions, e.g. detecting the most appropriate types of contents for a specific student and adapting the general information to her \cite{Fierrez-Aguilar2005_AdaptedMultimodal,Fierrez-Aguilar2005_BayesianAdaptation}.

Attention is defined as a conscious cognitive effort on a task \cite{tang2022using} and it's essential in learning tasks for a correct comprehension, since a sustained attention generally leads to better learning results. Nevertheless, the attention level measurement is not an easy task to perform, and it has been studied in depth in the state of art.

On the one hand, brain waves from an electroencephalogram (EEG) have demonstrated to be one of the most effective signals for attention level estimation \cite{chen2018effects,li2011real}. On the other hand, attention estimation through face videos obtained from a simple webcam (much less intrusive than measuring EEG signals) is also feasible but less effective in general. 

Now focusing in human behaviors related to the attention level that can be measured from face videos:

\begin{itemize}

\item The eye blink rate has been demonstrated to be related with the cognitive activity, and therefore the attention \cite{bagley1979effect, holland1972blinking}. Lower eye blink rates can be associated to high attention levels, while higher eye blink rates are related to low attention levels. ALEBk \cite{daza2021alebk} presented a feasibility study of Attention Level Estimation via Blink detection with results on the mEBAL database demonstrating that there is certain correlation between the attention levels and the eye blink rate. 

\item The student’s head pose can be also used to detect the visual attention which is a key learning factor \cite{luo2022three,zaletelj2017predicting,raca2015translating}. Also, the facial expression recognition has demonstrated to be a pointer in the state of human emotions and it’s highly related with the attention estimation in a learning environment \cite{monkaresi2016automated, mcdaniel2007facial, grafsgaard2013automatically}.

\item The student interest is also highly related with the attention levels, and some physical actions happen to be related with the interest. E.g., leaning closer to the screen is frequent when the displayed information is attractive or complex \cite{fujisawa2009estimation}.

\end{itemize}

Multimodal systems have the advantage of offering a global vision, taking into consideration several variables at different levels that affect the process of interest \cite{2018_INFFUS_MCSreview1_Fierrez}. As representative examples of multimodal attention level estimation we can mention: 1) \cite{zaletelj2017predicting} estimated attention levels using facial and body information obtained from a Kinect camera, and 2) \cite{peng2020predicting} used different features obtained from the face and also head movements (e.g., leaning closer to the screen).

In this context, the contributions of the present paper are:

\begin{itemize}

\item We perform an attention estimation study in a realistic e-learning environment including different facial features like pose, facial expressions, facial landmarks, eye blink information, etc. 

\item The study comprises different modules based on Convolutional Neural Networks trained to obtain facial features with potential correlation with attention. We evaluate each of these modules separately in the attention estimation task.

\item We propose a multimodal approach to improve the estimation level results in comparison with a monomodal system. The approach is based on a score-level fusion of the different face analysis modules. The results of our multimodal system are compared with another existing attention estimation system called ALEBk; both evaluated on the same mEBAL database. The results show that the proposed multimodal approach is able to reduce the error rates relatively by ca. $40\%$ in comparison with existing methods.

\end{itemize}

The rest of the paper is organized as follows: Section $2$ presents the materials and methods, including the databases and the proposed technologies to estimate attention levels. Section $3$ shows the experiments and results. Finally, remarks and future work are drawn in Section $4$.

\section{MATERIALS AND METHODS}
\subsection{Database: mEBAL}

The mEBAL database \cite{daza2020mebal} is selected for our study for several reasons. First, mEBAL is a public database which was captured using an e-learning platform called edBB in a realist e-learning environment \cite{hernandez2019edbb}. Second, mEBAL consists of $38$ sessions of e-learning (one per student) where the users perform different tasks of variable difficulty to create changes in the student’s cognitive load. These activities include finding the differences, crosswords, logical problems, etc. The sessions have a duration of $15$ to $30$ minutes. Third, mEBAL is a multimodal database with signals from multiple sensors including face video and electroencephalogram (EEG) data.

mEBAL includes the recordings of every session captured from $3$ cameras ($1$ RGB camera and $2$ NIR cameras). Besides, mEBAL used an EEG headset by NeuroSky to obtain the cognitive signals. Previous studies have also used this headset to capture EEG and attention signals \cite{rebolledo2009assessing,li2009towards,lin2018mental}. The EEG information provides effective attention level estimation \cite{chen2018effects,li2011real} because this information is sensitive to the mental effort and cognitive work, which varies significantly in different activities like learning, lying, perception, stress, etc. \cite{hall2020guyton, lin2018mental, chen2017assessing, li2009towards}.  

The resulting EEG information in mEBAL consists of $5$ signals in different frequency ranges. More precisely, the power spectrum density of $5$ electroencephalographic channels: $\delta$ ($<4$Hz), $\theta$ ($4$-$8$ Hz), $\alpha$ ($8$-$13$ Hz), $\beta$ ($13$-$30$ Hz), and $\gamma$ ($>30$ Hz) signals. From these channels, through the official SDK of NeuroSky, mEBAL includes information of the attention level, meditation level, and temporary sequence with the eye blink strength. The attention and meditation levels have a value from $0$ to $100$. The headset has a sampling rate of $1$ Hz. This work uses the attention level obtained from EEG headset as ground truth to train and evaluate our image-based attention level estimation system.

\begin{figure*}[t!]
\centering
\includegraphics[width=\textwidth]{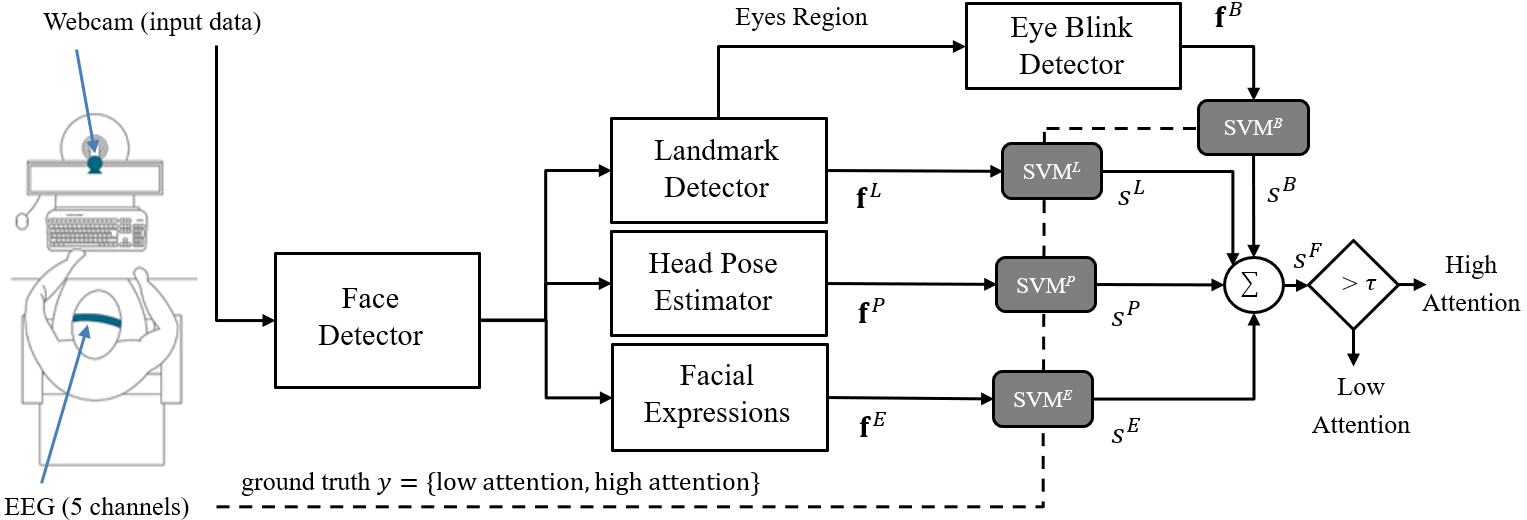} 
\caption{Block diagram of the multimodal attention level estimation approach. The dashed line represents ground truth used for training the SVMs.}
\label{Block_diagram}
\end{figure*}

\subsection{Face Analysis Modules}

Fig.~\ref{Block_diagram} shows the proposed multimodal approach for the estimation of the attention level. The framework includes the following modules:

\textbf{Face Detection Module:} First, we use a face detector to obtain $2$D face images. These images are used as input by the other technologies. The face position is estimated using the RetinaFace Detector \cite{deng2020retinaface}, a robust single-stage face detector trained using the Wider Face dataset \cite{yang2016wider}.

\textbf{Landmark Detection Module:} We use a landmark detector for two tasks. In the first one, we use landmarks for the localization of a region of interest for subsequent Eye Blink Detection. In the second one, we obtain the Eye Aspect Ratio (EAR) \cite{soukupova2016eye} for each eye (which is related to the opening of the eye), and the width and length of the nose and the head, as features to detect the head distance from the screen. We used the SAN landmark detector \cite{dong2018style}, which is a $68$-landmark detector based on VGG-$16$ plus $2$ convolution layers trained on a $300$-W dataset \cite{sagonas2016300}. Therefore, for each frame, $6$ features $\textbf{f}^L$ are used in our attention level estimation.

\textbf{Head Pose Estimation Module:} The head pose is estimated from $2$D face images. We obtain the vertical (pitch) and horizontal (yaw) angles to define a $3$D head pose from a $2$D image. The head pose is estimated using a Convolutional Neural Network (ConvNet) based on \cite{berral2021realheponet} trying to balance the speed and precision that maximizes the utility. This head pose is trained with Pointing $04$ \cite{gourier2004estimating} and Annotated Facial Landmarks in the Wild \cite{koestinger2011annotated} databases. So, for each frame we have $2$ angles as features $\textbf{f}^P$.

\textbf{Eye Blink Detection Module:} We used the architecture presented in ALEBk \cite{daza2021alebk} but trained on mEBAL from scratch, with only the eye blinks of the RGB camera. This detector has been inspired by the popular VGG16 Neural Network model and it’s a binary classifier (open or closed eyes) using two input images (cropped left eye and cropped right eye). For the region of interest localization, we also followed ALEBk: \textit{i)} face detection, \textit{ii)} landmark detection, \textit{iii)} face alignment, and \textit{iv)} eye cropping. For each frame, the module classifies into open or closed eyes, therefore for each frame we have a scalar value between $0$ and $1$ as a feature $\textbf{f}^B$.

\textbf{Facial Expression Detection Module:} Our model was inspired in the work~\cite{zhang2021learning}. The model was trained using FaceNet-Inception architecture pretrained with VGGFace2 and retrained with Google Facial Expression Comparison (FEC) dataset. The model follows the same experimental protocol proposed in~\cite{zhang2021learning}  to create a disentangled Facial Expression Embedding. The resulting Facial Expression Embedding $\textbf{f}^E$ consists of $16$ features that are used in our attention level estimation.

\subsection{Attention Level Estimation based on Facial Features}

The features $\{\textbf{f}^B,\textbf{f}^L,\textbf{f}^P,\textbf{f}^E\}$ obtained with the facial analysis models are used to estimate high and low attention periods. These features are obtained from very different behavior or physiological processes, that’s why we first analyze them separately, to understand the relation of each feature with the cognitive load estimation. To perform this analysis, we use information from mEBAL. On the one hand, we use the attention levels from the EEG band like ground truth for the experiments. On the other hand, we used RGB video recordings from the e-learning sessions in order to estimate the attention levels through image processing. 

The attention levels given by the band are captured every second ($1$Hz). In order to capture enough behavioral features, the attention level estimation proposed in this work is estimated according to one-minute time frames displaced every second (i.e., the attention is estimated every second based on the features captured during the last minute). We calculate the average band attention level per minute and concatenate the features vectors obtained when processing the images $\{\textbf{f}^B,\textbf{f}^L,\textbf{f}^P,\textbf{f}^E\}$, resulting in, for each attention level estimation (i.e., every second): $30$ frames per-second $\times$ $60$ seconds $\times$ features dimension.

The main goal of this work is to be able to detect one-minute time periods of high and low attention levels. Attention levels will vary on each student, that’s why we follow the same approach as ALEBk for the same mEBAL database, where two symmetric thresholds are proposed for high and low attention periods segmentation; high attention (attention higher than a threshold $\tau_H$), and low attention (attention lower than a threshold $\tau_L$) which are defined in relation to the probability density function (PDF) of the attention levels of all users.

\begin{figure*}[t]
\centering
\includegraphics[width=1\textwidth]{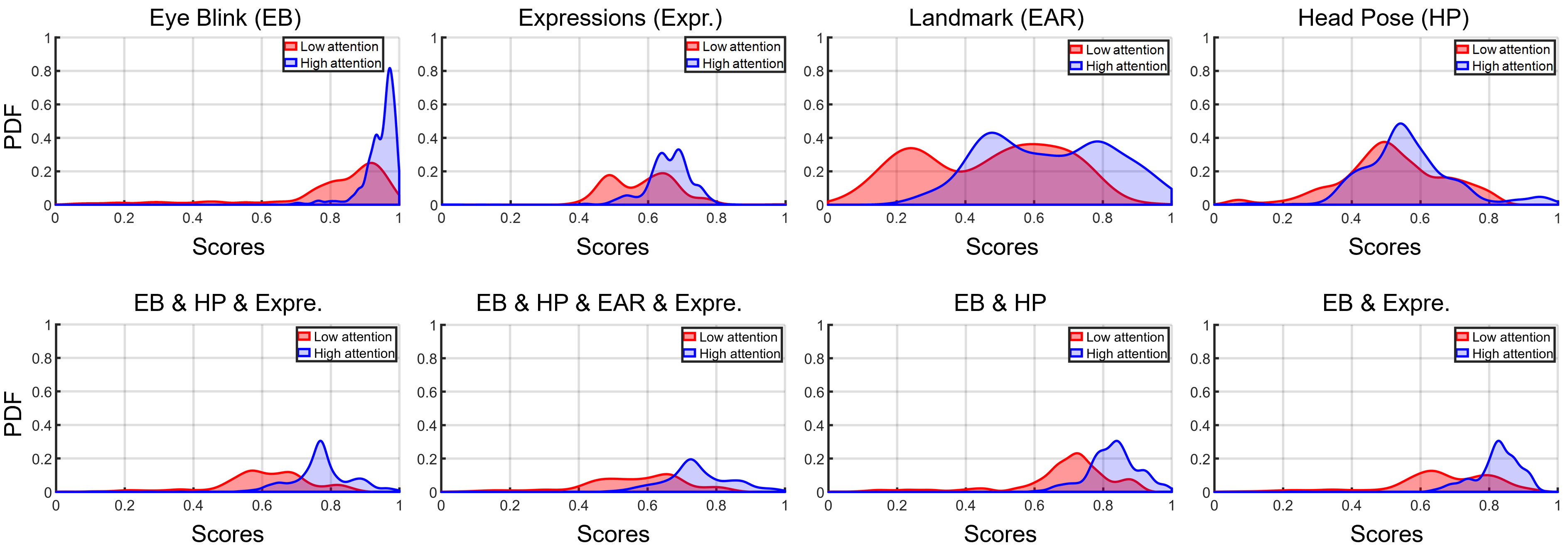} 
\caption{Probability density functions from scores obtained by our attention estimation systems during high/low attention (one-minute periods). Best systems/combinations for: monomodal systems in \emph{Top row} and multimodal systems in \emph{Bottom row}.}
\label{PDFs}
\end{figure*}

We individually evaluate eye blink ($\textbf{f}^B$), head pose ($\textbf{f}^P$), landmarks ($\textbf{f}^L$), and facial expressions ($\textbf{f}^E$) as binary monomodal classifiers (high or low attention). For all modules (see Fig.~\ref{Block_diagram}) we used the same classifying algorithm based on a Support Vector Machines (SVM) with a linear kernel using a squared I$2$ penalty with a regularization parameter between $1e^{-4}$ and $1e^{2}$, and a value of $1e^{-3}$ for the tolerance on stopping criterion.  

\textbf{Unimodal attention level estimation:} The training process is divided in the following steps: 1) Each frame was processed separately to obtain the four feature vectors (one from each module); 2) The feature vectors of all frames available in one minute video were concatenated ($30 $ frames/second $\times$ $60$ seconds $\times$ number of features of the module). 3) We trained one SVM for each module as binary classifiers of attention estimation per minute (high or low attention level). 

\textbf{Multimodal attention level estimation:} We combined the scores obtained for each of the facial analysis modules $\{s^B,s^L,s^P,s^E\}$   according to a weighted sum $s^F=\sum(w_Bs^B+w_Ls^L+w_Ps^P+W_Es^E)$, with equal weights in this initial study (subject to optimization in future work). Finally, the fused score $s^F$ was compared with the threshold $\tau$ to infer the attention level (high or low).

\begin{table}[t!]
\centering
\setlength\tabcolsep{7pt}
\renewcommand{\arraystretch}{1.1}
\caption{Attention estimation accuracy results using the mEBAL database for the proposed monomodal approaches. We set the value of $\tau_L$ on $10\%$ and $90\%$ for $\tau_H$. Also, two accuracy measurements were used, maximum accuracy, and equal error rate (EER) accuracy. }
\begin{tabular}{|l|c c|}
\hline
\bf{Module} & \bf{max acc} & \bf{acc=1-EER}   \\ 
\hline \hline
Eye Blink (EB) & 0.7639   & 0.7596  \\ \hline
Expressions (Expr.) & 0.6991   & 0.6705  \\ \hline
Landmark (EAR) & 0.6624 	& 0.6047 \\ \hline
Head Pose (HP) & 0.5977 	& 0.5793 \\ \hline
Landmark (Head Distance) & 0.5872 	& 0.5652 \\ \hline
\end{tabular}
\label{tab:ACC_ATT_Monomodal}
\end{table}

\section{EXPERIMENTS AND RESULTS}

\subsection{Experimental Protocol}

We used the protocol proposed in \cite{daza2021alebk}. The videos from the mEBAL database were processed to obtain the one-minute periods of high and low attention levels. We considered the lowest $10\%$ percentile for low attention, while high attention corresponds to the highest percentile ($\tau_L=10\%$ y $\tau_H=90\%$ values). In total, we obtained $3$,$706$ one-minute samples from the students in the database. From theses samples $1$,$852$ correspond to high  attention periods and $1$,$854$ to low attention periods.

We used the “leave-one-out” cross validation protocol, leaving one user out for testing and training using the remaining users; the process is repeated with all users in the database. The decision threshold is chosen at the point where False Positive and False Negative rates were equal and at the point where classification accuracy is maximized. 

\subsection{Unimodal Experiments}

Table \ref{tab:ACC_ATT_Monomodal} shows the accuracy results for each monomodal approach and top row in Fig.~\ref{PDFs} shows the probability density distributions of the obtained scores for each monomodal classifier. The results show that there’s a higher separability between distributions for modules based on eye blink (EB) and facial expressions (Expr.), obtaining maximum accuracy respectively of $76.39\%$ and $69.91\%$. As we can see, high attention levels (for most of the cases) are easier to recognize than low attention levels, which have a more spread density distribution. These results make sense in an evaluation environment like mEBAL, due to the fact that students would normally be focused with high attention moments during a short time tasks. Note that sessions in the mEBAL dataset have a duration between $15$ and $30$ minutes where the students solve different types of tasks.

As expected, the system based on head pose (HP) shows the second worst results. By itself this is not a clear attention estimation indicator, however, its information can be useful for multimodal approaches as we will see later. Landmarks features related to the proximity of students towards the camera (Head Distance), show two distributions almost entirely overlapped, indicating lack of utility for attention estimation. The features from this module were therefore removed from the multimodal approach. Regarding the Eye Aspect Ratio (EAR) feature estimated through landmarks, it provides worse results in comparison to the ones with eye blinks and facial expressions.

\begin{table}[t!]
\setlength\tabcolsep{7pt}
\renewcommand{\arraystretch}{1.1}
  \caption{Attention estimation accuracy results using the mEBAL database for the best combinations of the proposed multimodal approaches. We set the value of  $\tau_L$ on $10\%$ and $90\%$ for  $\tau_H$. Also, two accuracy measurements were used, maximum accuracy, and equal error rate (EER) accuracy. Systems: EB$=$ Eye Blink, Expr.$=$ Expressions, EAR$=$ EAR feature from Landmarks, HP$=$ Head Pose.}
  \label{tab:ACC_ATT_Multimodal}
 
  \begin{tabular}{|l| c c|}
    \hline
     \centering \bf{Modules}& \centering \bf{max acc} &  \bf{acc=1-EER} \\
    \hline \hline
    
     EB \& HP & 0.7969 & 0.7853 
  
    \\ \hline
     EB \& Expr.  & 0.7769 & 0.7734
   
    \\ \hline

      HP \& Expr. & 0.7590 &0.7334
     
    \\ \hline
     EB \& EAR & 0.7501 & 0.7377
  
    \\ \hline
     HP \& EAR  & 0.6764 & 0.6683
    \\ \hline  
     Expr. \& EAR & 0.6759 & 0.6486

  \\ \hline\hline\hline
     \textbf {EB} \& \textbf {HP} \& \textbf {Expr.} & \textbf {0.8215} & \textbf {0.8198}
 \\   \hline  
  
     EB \& HP \& EAR & 0.7947 & 0.7885 
 \\ \hline
      EB \& EAR \& Expr. & 0.7618 & 0.7407 
   \\ \hline
       HP \& EAR \& Expr. & 0.7250 & 0.7172
     \\ \hline\hline\hline
      EB \& HP \& EAR \& Expr. & 0.8066 & 0.7966 
     \\ \hline
  \end{tabular}
\end{table}

\subsection{Multimodal Experiments}

We now perform an analysis to detect which combination of the previously mentioned monomodal modules obtains the best results in our multimodal approach. The bottom of Fig.~\ref{PDFs} shows the probability density distributions of the best combination from the scores, and Table~\ref{tab:ACC_ATT_Multimodal} shows the accuracy for each multimodal approach. 

\begin{figure}[t!]
\centering
\includegraphics[width=1\columnwidth]{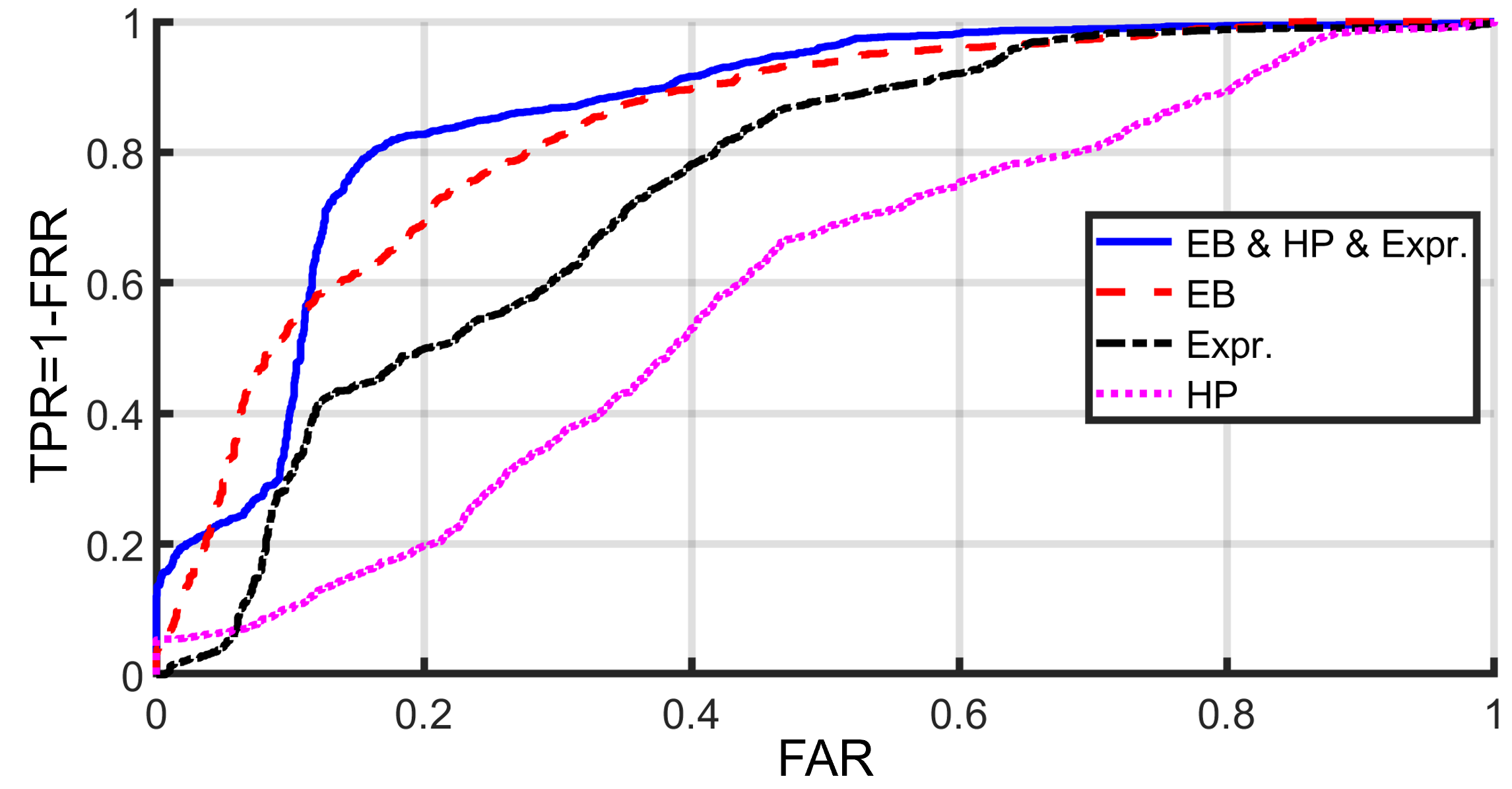} 
\caption{Comparison of the Receiver Operating Characteristic curve (ROC) obtained for the multimodal approach with the highest accuracy (blue line) and for each of the monomodal systems that belong to this combination.}
\label{ROC}
\end{figure}

The best scores are obtained from the combination of $3$ modules (eye blink detection, head pose, and facial expressions), see Fig.~\ref{ROC}, leading to an accuracy (1-EER) of $81.98\%$, which shows a significant improvement (ca. $25\%$ relative reduction in error rates) in relation to the best obtained result with our monomodal approach ($75.96\%$ 1-EER accuracy).

The second system with the best results is the combination of all 4 monomodal modules, which obtains a $79.66\%$ accuracy. This shows that including the system based on EAR worsens the results. 

The best combination using only two modules is the eye blink and head pose with $78.53\%$ accuracy, showing that the detection of the head pose can be of great help when it’s combined with other module to obtain more precise information about the context in the attention estimation; in fact, the head pose combination improves all combinations where it was included ($2$, $3$, or $4$ combinations). 

We finally compare with an existing approach: ALEBk \cite{daza2021alebk}, which recently proposed an attention classification method (low or high attention) based on the eye blink frequency per minute. The method was evaluated over the mEBAL dataset with a resulting best accuracy (1-EER) of $70\%$. The results shown on Table \ref{tab:ACC_ATT_Monomodal} and \ref{tab:ACC_ATT_Multimodal} (obtained by our monomodal and multimodal approaches, respectively) significantly outperform the ALEBK results over the same mEBAL database, same protocol, and same percentile of $10\%$ attention periods: from ALEBk best accuracy of $70\%$ to our best of ca. $82\%$, which means ca. $40\%$ relative reduction in error rates.

\section{CONCLUSION}

We performed an analysis of high and low attention estimation based on face analysis, using monomodal and multimodal approaches. We used different features that have proven to be effective for attention estimation, and for that, we have used recent technologies for Eye Blink Detection, Facial Expression Analysis, Head Pose, and Landmark Detection.

The results have showed the capacity of multimodal approaches to improve current methods for attention estimation. We have obtained ca. $82\%$ accuracy (as 1-EER) with a multimodal system that combines eye blink, facial expressions, and head pose features. In relation to the best obtained result with a monomodal system, we got ca. $76\%$ classification accuracy for the eye blink feature. Also, these results have corroborated a clear correlation between eye blink and attention. 

Our results have outperformed the ones obtained by ALEBk \cite{daza2021alebk}. The best obtained result by ALEBk was around $70\%$ classification accuracy, in comparison with our proposed multimodal system that obtained ca. $82\%$ accuracy. This means a relate improvement in error reduction (EER) of ca. $40\%$.

In future studies, we will explore other features that have shown a direct relation with attention levels like heart rate \cite{hernandez2020heart}, eye pupil size  \cite{rafiqi2015pupilware, krejtz2018eye}, gaze tracking \cite{wang2014eye}, keystroking \cite{2016_IEEEAccess_KBOC_Aythami,morales2016kboc}, etc. Also, we will explore new attention classifier architectures like Long and Short-Term Memory (LSTM Neural Networks), or other architectures, combining both short and long-term information. More advanced adaptive and user-dependent fusion schemes will be also studied \cite{2018_INFFUS_MCSreview1_Fierrez}.

\section{ACKNOWLEDGMENTS}

Support by projects: BIBECA (RTI2018-101248-B-I00 MINECO/FEDER), HumanCAIC (TED2021-131787B-I00 MICINN), TRESPASS-ETN (H2020-MSCA-ITN-2019-860813), and BIO-PROCTORING (GNOSS Program, Agreement Ministerio de Defensa-UAM-FUAM dated 29-03-2022). Roberto Daza is supported by a FPI fellowship from MINECO/FEDER.

\bibliography{aaai23}

\end{document}